\documentclass[runningheads]{llncs}

\usepackage{eccv}

\usepackage{eccvabbrv}
\usepackage{multirow}
\usepackage{overpic}
\usepackage{pgffor} 
\usepackage{tabularray}
\usepackage{graphicx}
\usepackage{booktabs}
\usepackage[table]{xcolor}
\usepackage{tikz}
\usetikzlibrary{positioning,fit,backgrounds}
\usepackage{wrapfig}
\usetikzlibrary{arrows.meta}
\usetikzlibrary{positioning}
\usepackage[accsupp]{axessibility}  
\usepackage{pifont}
\newcommand{\cmark}{\textcolor{green!70!black}{\ding{51}}}
\newcommand{\xmark}{\textcolor{red}{\ding{55}}}
\definecolor{grayrow}{gray}{0.95}
\usepackage{pgffor}

\newcommand{\safeincludegraphics}[2]{%
  \IfFileExists{#2}{%
    \includegraphics[width=#1]{#2}%
  }{%
    \fbox{\scriptsize N/A}%
  }%
}

\newcommand{\imgW}{0.16\linewidth}     
\newcommand{\protoW}{0.04\linewidth}   

\usepackage{hyperref}

\usepackage{orcidlink}

\begin{document}

\title{Bottom-up modeling of repeated elements \\ via single image analysis-by-synthesis} 

\titlerunning{Bottom-up modeling of repeated elements}

\author{Syrine Kalleli\inst{1}\thanks{Corresponding author: \texttt{syrine.kalleli@enpc.fr}}\orcidlink{0009-0003-5119-0596} \and
Alexei A. Efros\inst{2}\orcidlink{0000-0001-5720-8070} \and
Mathieu Aubry\inst{1}\orcidlink{0000-0002-3804-0193}}

\authorrunning{S.~Kalleli et al.}

\institute{LIGM, CNRS, Univ Gustave Eiffel, ENPC, Institut Polytechnique de Paris, France
\\
\and
UC Berkeley}

\maketitle
\begin{figure}[h!]
\centering
\vspace{-1em}
\resizebox{\columnwidth}{!}{%
\begin{tikzpicture}[
  font=\scriptsize,
  >={Latex[length=2.0mm,width=1.0mm]},
  title/.style={font=\scriptsize, align=center, text depth=0pt},
  steplbl/.style={font=\scriptsize, inner sep=1pt, fill=white},
  panel/.style={inner sep=0pt, outer sep=0pt}
]
\def\imgW{2.0cm}    
\def\protoW{1.0cm} 
\def\midgap{4mm}    
\def\outgap{7mm}   
\node[panel] (proto) {\includegraphics[width=\protoW]{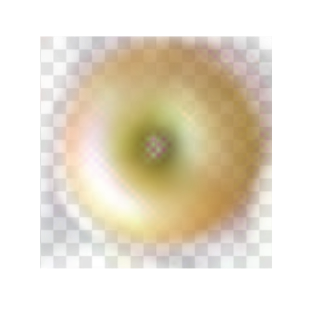}};
\node[panel, right=\midgap of proto] (s1) {\includegraphics[width=\imgW]{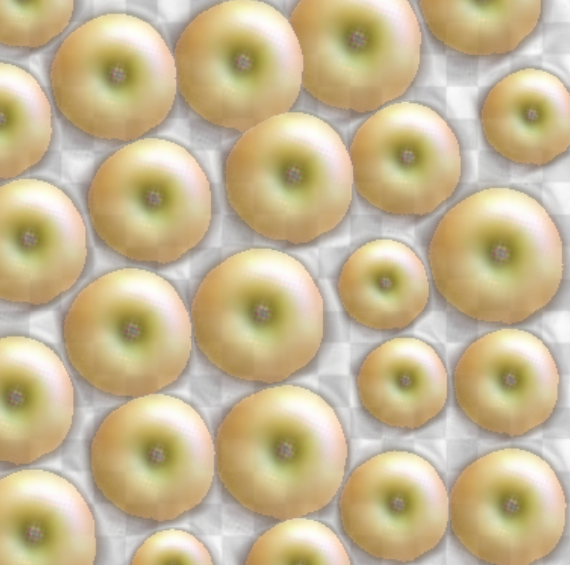}};
\node[panel, right=\midgap of s1]    (s2) {\includegraphics[width=\imgW]{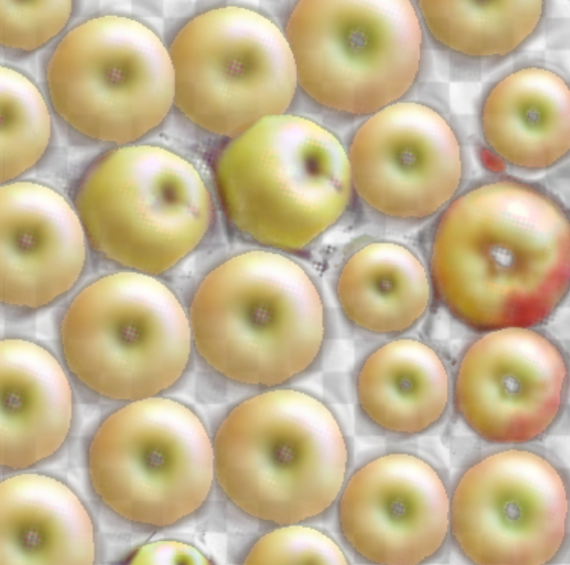}};
\node[panel, right=\midgap of s2]    (s3) {\includegraphics[width=\imgW]{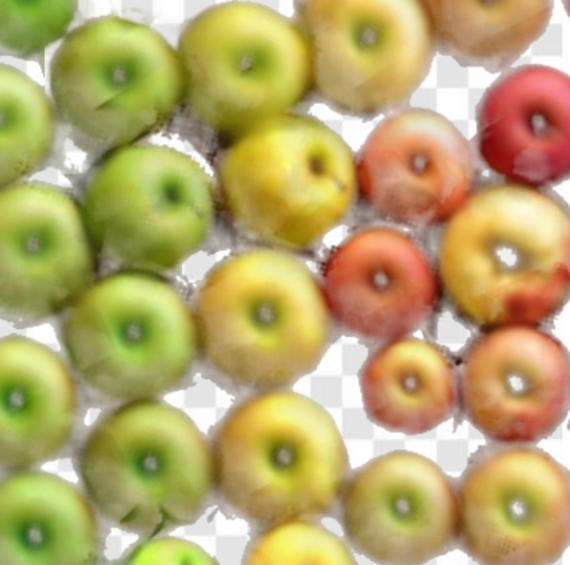}};
\node[title, above=-0.3mm of proto] (tproto) {Prototype};
\node[title, above=-0.3mm of s1] (ts1) {Position};
\node[title, above=-0.3mm of s2] (ts2) {+Instance App.};
\node[title, above=-0.3mm of s3] (ts3) {+Color};
\begin{scope}[on background layer]
  \node[draw, dashed, rounded corners, inner sep=0.6mm,
        fit=(proto)(s3)(tproto)(ts3)] (mid) {};
\end{scope}
\node[panel, anchor=east, xshift=-\outgap] (input) at (mid.west |- s1)
      {\includegraphics[width=\imgW]{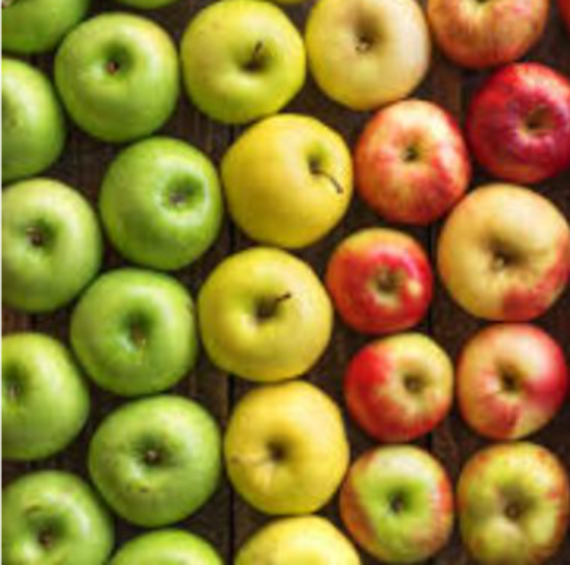}};
\node[title, above=-0.3mm of input] {Input};
\node[panel, anchor=west, xshift=\outgap] (recon) at (mid.east |- s1)
      {\includegraphics[width=\imgW]{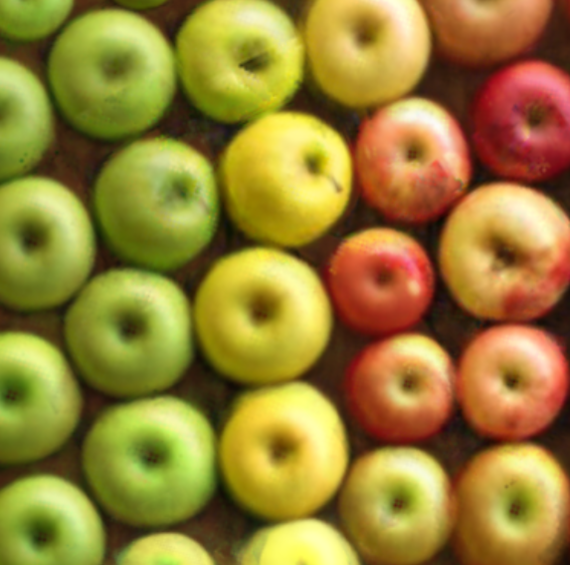}};
\node[title, above=-0.3mm of recon] {Reconstruction};
\def\trim{0.2mm}  
\draw[->, line width=0.8pt] ([xshift=-\trim]proto.east) -- ([xshift=\trim]s1.west);
\draw[->, line width=0.8pt] ([xshift=-\trim]s1.east) -- ([xshift=\trim]s2.west);
\draw[->, line width=0.8pt] ([xshift=-\trim]s2.east) -- ([xshift=\trim]s3.west);
\draw[->, line width=0.8pt] ([xshift=-0.8]input.east) -- ([xshift=1]mid.west |- s1.west);
\draw[->, line width=0.8pt] ([xshift=-0.8]mid.east |- s1.west) -- ([xshift=1]recon.west)
      node[steplbl, fill=none, midway, above, yshift=1pt] {+bkg};
\end{tikzpicture}%
}
\caption{\textbf{Element and image models}. Our goal is to learn a single image-space prototype that captures the common appearance of repeated elements within one image. Our method reconstructs the input image by transforming a shared prototype with instance-specific pose and appearance parameters. The resulting foreground is composited with a low-resolution background to produce the final reconstruction.}
\label{fig:overview}
\vspace{-1em}
\end{figure}
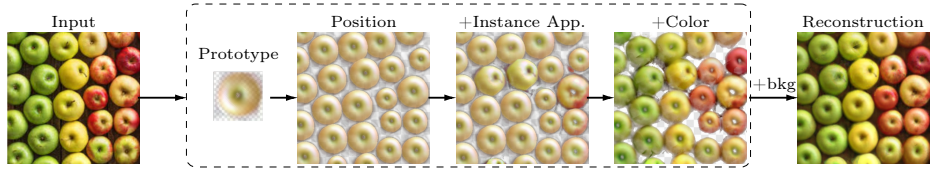

\begin{abstract}
    We address the problem of discovering repeated elements from a single image. In contrast to existing approaches that depend on large annotated datasets, curated multi-image collections, or object segmentation masks, we show that a single image can suffice to learn a meaningful object model in a completely bottom-up fashion, without any prior knowledge beyond a coarse scale prior. Our method learns a tunable image-space prototype of the repeated elements through a reconstruction objective, enabling the model to identify and synthesize consistent object instances within the same image.
    Experiments on 116 real images from the FSC-147 dataset demonstrate that our method successfully learns coherent element models and captures intra-category variation on challenging images. Qualitative results reveal superior reconstructions and interpretable decompositions compared to classical decomposition, joint alignment, and 3D object modeling methods, while maintaining a simple 2D formulation. These results suggest that meaningful object discovery can emerge from single-image learning alone.
\end{abstract}

\section{Introduction}






Pattern recognition -- discovering approximately repeating elements within the sensory data -- is one of the most fundamental building blocks of animal and human intelligence. It is what allows organisms to start building representations of their sensory world. 

Bottom-up pattern recognition has been a bedrock of early computer vision, including unsupervised clustering, bottom-up segmentation, texton modeling, and unsupervised object discovery in image collections. In contrast, modern computer vision has largely forgone bottom-up methods in favor of top-down supervised approaches (supervised segmentation, supervised object counting, \etc).  Such methods perform very well on benchmarks but often fail when presented with novel, out-of-distribution data. Moreover, their reasoning is often difficult to interpret, resulting in a black-box effect, which is problematic, in particular, for scientific applications. 

In this paper, our goal is to revisit the bottom-up tradition of modeling visual data, starting from pixels, without supervision or annotation beyond a coarse scale prior. We additionally find that this object-scale prior matters only for the smallest objects. Our task is a special case of the pattern recognition problem -- identifying and modeling repeated elements within a single image. We take the analysis-by-synthesis approach, using a reconstruction objective to jointly optimize a parametric model of the repeated element, together with each instance's occurrence and parameters. Our resulting model is not only effective compared to previous works, but is also highly explainable, as visualized in Figure~\ref{fig:overview}.

From the technical point of view, our architecture is inspired by deep object-centric image decomposition methods, and in particular capsule networks~\cite{sabour2017dynamic,kosiorek2019stacked} and sprite-based methods~\cite{smirnov2021marionette,monnier2021dtisprites}. However, their setting is quite different from ours -- they assume access to multiple images of multiple objects in different configurations. More importantly, they struggle to work beyond simple synthetic data, except in much simpler settings, such as a single object per image, where they become similar to congealing approaches. 
Critical to the success of our method are a low-dimensional parametric element generation module, which enables the modeling of slight element variations without sacrificing interpretability and controllability, and a crop-based curriculum learning strategy. 


 We demonstrate our results on a curated subset of the FSC-147 counting dataset~\cite{ranjan2021learning} of real images, assessing not only counting but also the modeling of elements and images. 

\section{Related Work}
In this section, we give an overview of methods that focus on discovering and modeling single-image and multiple-image repetitions. While our work falls into the single-image category, it also builds on ideas that have been developed for multiple images, from which we also build most of our baselines. We summarize distinctions with related work in Table~\ref{tab:rw}.
\begin{table}[t]
\centering
\caption{Comparison of our method with related approaches.}
\label{tab:rw}
\resizebox{\columnwidth}{!}{%
\begin{tabular}{lccc}
\toprule
Method & Object discovery & Object model & No regularity assumption  \\
\midrule
Single-Image Generative Models~\cite{shaham2019singan, wang2025sindiffusion, kulikov_sinddm_2022, shocher_ingan_2019} & \xmark & \xmark & \cmark \\
Regular pattern discovery~\cite{muller2007image, schaffalitzky1999geometric, hays2006discovering, Liu_2015_ICCV, doubek2010image} & \cmark & \xmark & \xmark \\
3D inverse rendering~\cite{gadelha20173d, zhang2023seeing, cheng2024structure} & \xmark & \cmark & \cmark \\
\textbf{Ours} & \cmark & \cmark & \cmark \\
\bottomrule
\end{tabular}
}
\end{table}
\subsection{Repeated elements in a Single Image}
Using repetition to detect and model important image elements is an idea that dates back to early work on visual texture. 
Early methods often attempt to explicitly model repeated elements but rely on strong geometric assumptions. For example, \cite{leung1996detecting} assumes that the elements are close to each other, while \cite{schaffalitzky1999geometric} assumes that the elements lie on a common plane. Other approaches focus on texel discovery~\cite{ahuja2007extracting} through feature matching or exploit structured regularity, such as 2D lattices~\cite{hays2006discovering,Liu_2015_ICCV,doubek2010image} and 1D linear patterns~\cite{muller2007image}.

Identified recurrent elements were often used for image processing purposes, such as editing~\cite{cheng2010repfinder}, super-resolution~\cite{glasner2009super, huang2015single}, texture synthesis~\cite{efros1999texture} and novel-view synthesis~\cite{violante2025splat}. 
\cite{chen2022learning} revisits this line of work and performs near periodic patterns image completion by optimizing a continuous MLP-based implicit representation of repeated patterns.
A related line of work learns a generative model of a scene from a single image by exploiting its internal patch statistics, using GANs~\cite{shaham2019singan,shocher_ingan_2019} to match multi-scale patch statistics or diffusion models~\cite{wang2025sindiffusion, kulikov_sinddm_2022} for higher-quality, more diverse generation. These methods capture recurrence only implicitly, as a distribution over patches. In contrast, we learn an explicit, parametric model of the repeated element.
More recently, several works use repetition for 3D modeling, 
learning from exact duplicates~\cite{cheng2024structure} or category-specific instances~\cite{gadelha20173d, zhang2023seeing, 
sinlayout2024}. 
Similarly to our work, ~\cite{zhang2023seeing} learn a parametric object model of the repeated object. 
However, these 3D methods are expensive and require clean ground-truth segmentation masks. Furthermore, they struggle with 2D scenarios, such as aerial imagery, or cases where the lack of diverse viewpoints prevents robust 3D reconstruction.\\

Moving away from attempts at modeling, counting methods focus on estimating the number of repeated elements, and often identifying their location. Counting methods can be broadly categorized depending on the type of supervision they require. Early approaches were fully supervised, and limited to a single object class of interest~\cite{lempitsky2010learning}. Class-agnostic counting, introduced in~\cite{lu2018class}, reformulates the problem as locating repetition of a given exemplar, enabling generalization to unseen classes at test time.
Many recent works~\cite{liu2022countr, knobel2024learning, djukic2023low} leverage transformers to predict density maps. In particular,~\cite{knobel2024learning} generate synthetic training data by copy-pasting objects within the same feature cluster defined by DINO representations.
With the extension of one of the main datasets of the field, FSC-147~\cite{ranjan2021learning}, to few-shot counting and detection (FSCD)~\cite{nguyen2022few}, several follow-up methods adopt detection-based formulations, including C-DETR~\cite{nguyen2022few}, Dave~\cite{pelhan2024dave}, PseCo~\cite{huang2024point}, GeCo~\cite{pelhan2024novel} and TMR~\cite{jo2025tmr}. For our baselines and ablations, we use TMR~\cite{jo2025tmr} which provides state-of-the-art results, can segment patterns that are not associated with object categories, and provides as output a count, bounding boxes, and optionally segmentations. 
These approaches can be further adapted to zero-shot counting, using a textual description of the object instead of an exemplar~\cite{amini2023open}. Some methods enable having both text queries or exemplars as input~\cite{amini2024countgd,huang2024point}. Others tackle the more challenging exemplar-free setting, such as RepRPN~\cite{ranjan2022exemplar}. Recently, the large-scale plant counting dataset TPC-268~\cite{xu2026plant} was released, providing fine-grained and biodiverse images.
Going further, ABC123~\cite{hobley2024abc} proposes multi-class, exemplar-free counting through pretraining on labeled synthetic data. Since it does not require exemplars or text prompts, we argue that it is the most relevant baseline for our unsupervised approach in terms of counting results. However, unlike our method, it does not provide any model of the repeated element or the image.

\subsection{Repeated elements in Image Collections}
\paragraph{Single element type.} 
Approaches to align images to a shared canonical template can be traced back to congealing~\cite{learned2005data, huang2007unsupervised} where an image set is iteratively aligned to a shared image through entropy minimization. 
More recent works rely on Spatial Transformer Networks~\cite{jaderberg2015spatial} (STN) to differentiably warp instances.
Some use the latent space of GANs~\cite{peebles2022gan, mu2022coordgan} to learn a model across a large dataset depicting a single object class.
Others leverage the rich representations of self-supervised models such as DINO~\cite{caron2021emerging}, which provide powerful cues for semantic correspondence~\cite{Amir:2021:VITdescriptoers,ofri2023neural,gupta2023asic,barel2025spacejamlightweightregularizationfreemethod,hirsch2025fastjam}. 
In particular, SpaceJAM~\cite{barel2025spacejamlightweightregularizationfreemethod} proposes a fast, lightweight, joint alignment with an implicit atlas. 
Most recently, FastJAM~\cite{hirsch2025fastjam} accelerates convergence by training a graph neural network on sparse keypoints from a supervised matcher. 
Similar to these works, our discovered instances are implicitly aligned. However, our approach does not rely on access to a collection of images of a single object, which provides strong implicit supervision.

\paragraph{Multiple element types.} 
Object-centric image decomposition methods~\cite{villa2024unsupervised}, aim at discovering structured, object-centric decompositions of images. While these works successfully handle multiple objects, they are generally evaluated on controlled, synthetic datasets~\cite{karazija2021clevrtex}. Methods such as AIR~\cite{eslami2016attend} and SPAIR~\cite{crawford2019spair}, factorize scenes into explicit latent variables for "what", "where", and "presence." These methods attend to local regions or "glimpses" and typically rely on a costly sequential inference. To improve scalability, SPACE~\cite{lin2020space} introduced a parallelized patch-based decomposition. Other works, such as MoNet~\cite{burgess2019monet} and IODINE~\cite{ greff2019multi} compute full-image pixel-level masks to isolate objects. However, these models suffer from high computational overhead and struggle to scale due to their reliance on iterative refinement or sequential attention. Slot Attention~\cite{locatello2020object} and follow-ups~\cite{singh2021illiterate,seitzer2022bridging, biza2023invariant, jia2023improving} improve this by using a single step of iterative slot attention. 
Closest to our work are sprite-based methods~\cite{monnier2021dtisprites, smirnov2021marionette}, which learn an image-space dictionary of objects that can be transformed and placed onto a canvas. This paradigm is particularly attractive for our task since it compresses the scene into a set of controllable, learnable objects. However, DTI-Sprites~\cite{monnier2021dtisprites} is not designed to handle a large number of objects or a single-image setup, and MarioNette~\cite{smirnov2021marionette} is patch-based and is limited by its grid resolution. Additionally, in both of these prior works, the prototype corresponding to a single object is fixed. In contrast, our approach allows for appearance variations for each instance beyond those captured by colorimetric and simple geometric transformations, which enables it to handle real images.

\section{Method}


\begin{figure}[t]
    \centering
    \includegraphics[width=\linewidth]{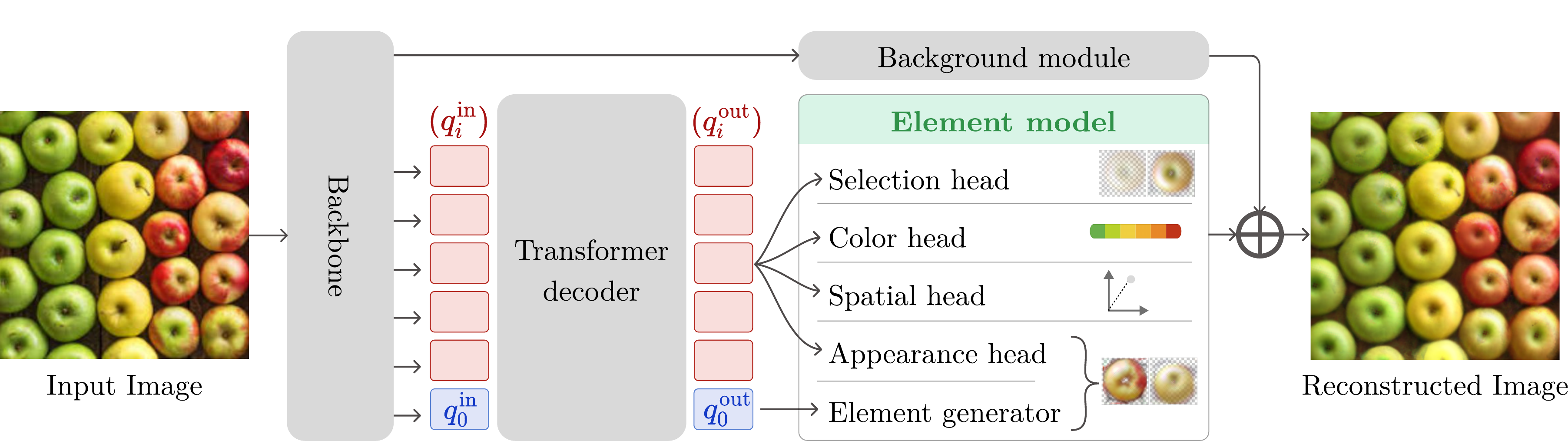}
    \caption{ \textbf{Architecture overview.} The input image is encoded into patch features and processed by two parallel branches: a background module for low-resolution predictions and an element prediction module. The latter uses a transformer decoder with an appearance query $q_0^{in}$ and instance queries $\{q_i^{in}\}_{i=1}^N$ to produce output tokens $\{q_i^{out}\}_{i=0}^N$. While the $q_0^{out}$ captures shared features, each $q_i^{out}$  independently predicts existence probability, spatial parameters (position, rotation, and scale), a color transform, and an appearance code for element generation. Finally, these generated elements are transformed and composited onto the background via alpha blending, weighted by their existence probabilities.} 
    \label{fig:method}
\end{figure}

In this section we first present the core of our approach, namely in Section~\ref{sec:model} our element-based image model and the associated architecture, and in Section~\ref{sec:unsupervised} our single-image fully unsupervised optimization. 
Then, in Section~\ref{sec:supervised}, we describe how to incorporate annotations as exemplar bounding boxes or element locations such as those provided by counting models. Finally, in Section~\ref{sec:pretraining}, we present a synthetic pretraining that significantly speeds up single-image optimization and improves our results. 

\subsection{Element-based modeling and architecture} \label{sec:model}

\paragraph{Overview.}
Given a single input image $I \in \mathbb{R}^{h \times w \times 3}$ 
containing multiple repeated instances of the same element, our objective is to jointly learn an element model and discover its instantiations in the scene. We represent the scene as the composition of a background image $I_B \in \mathbb{R}^{h \times w \times 3}$ and $L$ foreground layers. Each foreground layer $l_i$ corresponds to one element instance and is obtained by applying a predicted spatial and color transformation to a shared low-dimensional
element model. 
An overview of our image modeling approach is presented in Figure~\ref{fig:overview} and an overview of our architecture can be found in Figure~\ref{fig:method}.

\paragraph{General architecture.} More specifically, the input image is first processed by a backbone into patch features. These features are then fed into two parallel networks: a background module, which predicts a low-resolution background image, and an element prediction module, which models all foreground elements. 

The element prediction module starts with a transformer decoder, which takes as input a set of learnable instance queries $\{q^{in}_i\}_{i=1}^N$ and a single global appearance query $q^{in}_0$ and cross-attends to the image features. It produces output tokens $\{q^{out}_i\}_{i=0}^N$. Intuitively, the goal of the global appearance token $q^{out}_0$ is to capture the general element appearance, while the instance queries $\{q^{out}_i\}_{i=1}^N$ aim at learning element occurrence and instance-specific pose and appearance.

Each output instance token $\{q^{out}_i\}_{i=1}^N$ is processed {independently} by four specialized heads: (a) a selection head that predicts an existence probability; (b) a spatial head that predicts scale, rotation, and translation; (c) a color head that estimates a linear color transformation in RGB space; and (d) an appearance head, that predicts a low-dimensional appearance code (1D in all experiments), which conditions the element generation module, described in the next paragraph, to produce instance-specific elements. The instance-specific elements produced by the element generation module and the transformation parameters (b-c) are then used to model each instance. Finally, all generated instances are composited through alpha-blending, with opacity weighted by their probability (a), onto the predicted background to produce the reconstructed scene.

\paragraph{Element model.} 
The element generation module can be thought of as extending the element model beyond a single element prototype to a low-dimensional (1D in our case) parametric family of appearances. Its goal is to model small variations in appearance across instances and 3D effects.
The RGBA image of a specific element $i\in \{1,...,N\}$,  $I_i \in \mathbb{R}^{w_P \times h_P \times 4}$, is generated by a small generator network from the concatenation of the global appearance token $\{q^{out}_0\}$ and the instance appearance tokens  $\{q^{out}_i\}$ are  predicted by the appearance head. 

\paragraph{Architecture details.}
The backbone is a frozen pre-trained ViT-B/16 DINOv3~\cite{simeoni2025dinov3}  model, selected for its rich self-supervised object-centric representations, whose features are projected to $64$ dimensions by a learnable linear layer.
The background module applies global average pooling followed by a 2-layer MLP with 64 hidden neurons predicting a background color. 
The element prediction module uses $60$ instance queries and one global appearance query, all of dimension $64$. It consists of 3 standard cross-attention/self-attention blocks. %
Each of the selection, color, spatial, and appearance heads is implemented as a 2-layer MLP with 64 hidden neurons. The element generation module is also a 2-layer MLP with 64 hidden neurons.

The prototype resolution is $(w_P, h_P) = (64, 64)$. We use images of size $(224,224)$ 
 but only reconstruct them at $(128, 128)$ resolution to limit computational cost.
The background and the object layers are composited via alpha blending in a fixed order. The spatial module predicts the \textit{inverse} of the transform fed to the STN~\cite{jaderberg2015spatial} as we found it leads to much more stable training.

\subsection{Unsupervised single-image optimization} 
\label{sec:unsupervised}
\paragraph{Loss.}
Our unsupervised loss $\mathcal{L_\text{unsup}}$ is the combination of a reconstruction loss $\mathcal{L}_{\text{rec}}$ and a sparsity regularization loss  $\mathcal{L}_{\text{reg}}$ which prevent the element module from modeling background regions:
\begin{equation}
\mathcal{L_\text{unsup}}
=
\mathcal{L}_{\text{rec}}
+
\lambda_{\text{reg}} \mathcal{L}_{\text{reg}},
\end{equation}
where $\lambda_{\text{reg}}$ is a scalar hyperparameter.
The reconstruction loss combines a pixel-wise Huber loss and a structural similarity term:
\begin{equation}
\mathcal{L}_{\text{rec}}
=
\mathcal{L}_{\text{Huber}}
+
\lambda_{\text{SSIM}} \bigl(1-\mathrm{SSIM}(I,\hat{I})\bigr),
\end{equation}
where $I$ and $\hat{I}$ denote the target and reconstructed images,  
and the Huber loss is defined as
\begin{equation}
\mathcal{L}_{\text{Huber}}
=
\frac{1}{N_\text{pix}}\sum_{x}
\begin{cases}
\frac{1}{2}(I_x-\hat{I}_x)^2 & \text{if } |I_x-\hat{I}_x|\le\delta,\\
\delta\left(|I_x-\hat{I}_x|-\frac{\delta}{2}\right) & \text{otherwise},
\end{cases}
\end{equation}
where the sum is over pixels $x$ in the image, $N_\text{pix}$ is the total number of pixels, $I_x$ and $\hat{I}_x$ are the values of the target and reconstructed images at pixel $x$, and $\delta>0$ is a scalar hyper-parameter.

The regularization loss penalizes the number of elements used for the reconstruction:
\begin{equation}
    \mathcal{L}_{\text{reg}}=\frac{1}{K}\sum_{k=1}^{K} |p_k|,
\end{equation}
where $K$ is the number of element tokens and $p_k$ is the probability predicted by the selection head for the $k$-th element token.

\paragraph{Curriculum learning.} Jointly training all modules from the start can cause ambiguities (\eg the element generation module can learn to move, scale, and color the elements or reconstruct the background). To avoid this, we adopt a 3-step curriculum learning strategy. First, we train the background model 
keeping all other modules frozen. Second, we train all other components, except the appearance prediction head, whose output we fix at zero to enforce a shared appearance across all instances. Third, we unfreeze the appearance prediction head, which allows for instance-specific adaptation of the element appearance.

\paragraph{Crop-based optimization.} Single-image optimization leads 
to uninteresting solutions, such as optimizing colored blobs to reconstruct the input image. To avoid this, we apply strong spatial data augmentation as a form of regularization. For each input image, we generate fixed-size training crops using random cropping and rotation.
This augmentation is such that the smallest side of the object is between $10\%$ and $50\%$ of the crop side. We obtain the scale prior needed to define the crop size range for each image from the exemplar instance annotations available in the dataset.  



\paragraph{Full image reconstruction.} Since our model is trained on crops, reconstructing the full image is non-trivial: its scale and number of elements differ from the crops seen during training. We therefore extract overlapping crops with a sliding window and predict element parameters and existence probabilities per crop, which may model the same element multiple times. To remove duplicates, we map predicted transformations to global image coordinates, discard elements below a probability threshold $\delta_{\text{prob}}$, and select the remaining ones sequentially in order of decreasing probability: a prediction is kept only if it reduces the reconstruction loss by more than $\delta_{\text{loss}}$ and its overlap with previously selected elements stays below $\delta_{\text{overlap}}$. 
Finally, we run a full-image refinement stage, jointly optimizing the selected elements' parameters, the average predicted background, and the prototype generation to improve reconstruction quality.

\paragraph{Optimization details.} In our experiments, we use $\lambda_{\text{SSIM}}=0.01$, $\lambda_{\text{reg}}=0.001$, and $\delta=1$ in the Huber loss. We train our network on image crops using the Adam optimizer with a learning rate of $10^{-3}$, a linear warmup of 500 iterations, and a batch size of 64. 
To speed up training, we initialize the prototype generator with a gaussian blob. We use $100$ iterations, $500$ iterations, and $2500$ iterations for the three steps of the curriculum training, respectively.
 For the full image reconstruction, we use a sliding window of average crop size, a stride of $25\%$, a probability threshold $\delta_{\text{prob}}=0.5$, a reconstruction threshold $\delta_{loss}=10^{-5}$, and an overlap threshold $\delta_{\text{overlap}} =0.1$. The final refinement is performed for 200 iterations with a learning rate of $10^{-4}$.

\subsection{Optional annotation inputs} \label{sec:supervised}

Our method can of course benefit from additional inputs. We consider two types of input: (i) the bounding box of an element, which is typically provided by counting datasets and can disambiguate element definition; (ii) location supervision, in the form of a set of points corresponding to all the elements' centers, which is a typical output for counting methods.  These are particularly beneficial for cases with part-whole ambiguity, e.g., between a pair of glasses and its half. These are often impossible to disambiguate from purely visual cues, in particular, in the common cases of regular arrangements and symmetric objects.

\paragraph{Exemplar supervision.} 
To take advantage of exemplar bounding boxes, we add for each available exemplar box region $R$ an additional loss, $\mathcal{L}_{\text{ex}}$. It is defined by selecting the predicted element which best fits the box and encouraging it to reconstruct the exemplar region when combined with only the background. 
Let us assume that for a target image $I \in \mathbb{R}^{h\times w\times3}$ the model predicts a background image 
and 
$N$ elements with probabilities
$p^n \in [0,1]$, spatially transformed masks $M^n \in [0,1]^{h\times w}$, and colored appearances
$A^n \in \mathbb{R}^{h\times w\times3}$. 
We first select the best-fitting instance $n^*$ with the exemplar box $R$ as 
\begin{equation}
n^* = \displaystyle \arg\max_{n \in \{1,\dots,N\}}
\sum_{x\in R} p^n M^n_x,
\end{equation}
where the sum is over the pixels $x$ in $R$ and $M^n_x$ is the value of the mask $M^n$ at pixel $x$. We then compute a single-exemplar reconstruction of the image with $\hat{I}_{n^*}$ by alpha-blending the selected element appearance $A_{n^*}$ with the background, using the mask $p_{n^*}M_{n^*}$. 
The exemplar loss $\mathcal{L}_{\text{ex}}$ is then defined to encourage $\hat{I}_{n^*}$ to be close to the target $I$ in the annotated exemplar bounding box region $R$ and the probability $p_{n^*}$ to be close to 1:
\begin{equation}
\mathcal{L}_{\text{ex}} =
\mathcal{L}_{\text{rec | R}}\big(
I,
\hat I_{n^*}\big)
+ \lambda_{\text{pres}} (1 - p_{n^*})
\label{eq:exemplar_loss}
\end{equation}
where  $\mathcal{L}_{\text{rec | R}}$ is the reconstruction loss defined in the previous section restricted to the region defined by $R$, and $\lambda_{\text{pres}}$ is a scalar hyperparameter set to $\lambda_{\text{pres}}=0.1$ in our experiments.

\paragraph{Location supervision.}
In images with complex contrasted backgrounds, our unsupervised approach might use colored elements to approximate the background. This could be solved by using a detection-based counting network, trained on large-scale data, to produce a set of pseudo ground-truth element location annotations. 
In that case, we use the Hungarian algorithm to find an optimal one-to-one assignment between the instance locations predicted by the element prediction module and the pseudo-labels provided by the counting model. With this optimal matching, we define a loss:
\begin{equation}
    \mathcal{L}_{loc}=\lambda_{pos}\mathcal{L}_{\text{pos}} + \lambda_{prob+} \mathcal{L}^{\text{prob+}}+ \lambda_{prob-} \mathcal{L}^{\text{prob-}},
\end{equation}
where $\mathcal{L}_{\text{pos}}$ is the sum squared distance between pseudo-labels and their matched predictions, $\lambda_{pos}$, $\lambda_{prob+}$ and $\lambda_{prob-}$ are scalar hyperparameters, and $\mathcal{L}^{\text{prob+}}$ and $\mathcal{L}^{\text{prob-}}$ are losses on the probabilities similar to the ones of DETR~\cite{carion2020end} which encourage matched elements to have high probabilities and non-matched elements to have low probabilities. In our experiments, we use $\lambda_{pos}=0.5$, $\lambda_{prob+}=0.5$ and $\lambda_{prob-}=0.2$ and remove the sparsity regularization loss $\mathcal{L}_\text{reg}$ when using location supervision.


\subsection{Synthetic pretraining} \label{sec:pretraining}
Optimizing a per-image model from scratch is computationally inefficient, and most of the optimization time is spent finding coarse object locations with simple blob objects, and the optimization sometimes ends up in bad local minima. 
Thus, in order to speed up the optimization and improve results, we pretrain the network on many synthetic images with repeated elements. Our architecture and training objectives remain the same, but the task is significantly more difficult since the element generation module now needs to learn a general element model. Thus, this pretraining provides very poor image and object models when applied directly to a real image but a good initialization for our single-image optimization.




\paragraph{Synthetic Data Generation} 
Pretraining our network with real images would be appealing, but no large scale dataset of images with repeated objects is available, and real data is often too difficult, leading the model to reconstruct the scene with colored blobs.  Synthetic data provides an appealing alternative as the difficulty can be controlled, unlimited training examples can be generated, and the model can be supervised without cost. We leverage images and segmentations from COCO~\cite{lin2014microsoft} to generate our synthetic images. To generate a synthetic image, we first choose randomly a segmented object, and a background. The background can be a uniform color or a heavily blurred crop of an image from the dataset. The foreground is constructed by pasting copies of the selected object several times with random position, scale, rotation, and color jitter. In practice, we pretrain on 5,000 synthetic images. Examples can be seen in our supplementary material. 

\paragraph{Pretraining details.}  The pretraining is done in a supervised fashion with the reconstruction, exemplar, and location supervision objectives. We pretrain the network on 5,000 synthetic images with the Adam optimizer and a learning rate of $10^{-3}$ starting with 500k iterations using only images with uniform background, then with an equal mix of uniform and blurred background for 500k iterations. 
  

\section{Experiments}\subsection{Dataset and Evaluation Protocol}
\begin{figure}[t]
    \centering
    \includegraphics[width=\linewidth]{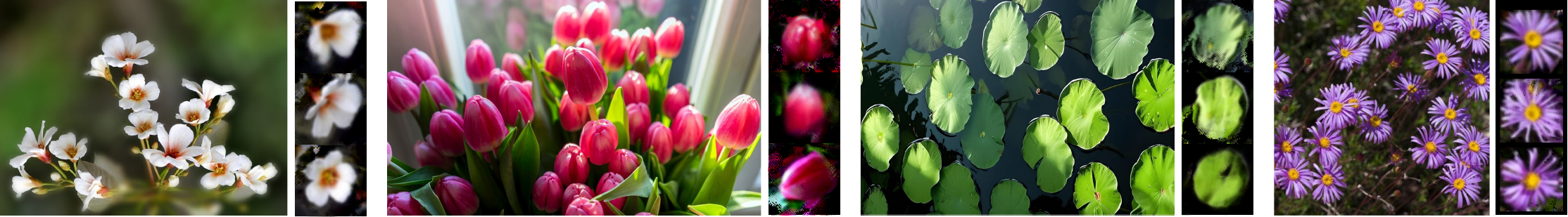}
    \caption{Samples from TPC-268~\cite{xu2026plant} (left) and our learned prototypes (right). }
    \label{fig:tpc}
\end{figure}
\subsubsection{Dataset.} To evaluate and compare methods for our task, we build a dataset by curating images from the 29 test categories of the standard FSC-147~\cite{ranjan2021learning} counting dataset. 
For each category, we select four images to maintain a balanced representation of classes. 
We favor images with simpler backgrounds and diverse object poses and counts. This results in 116 images, annotated with object count and three exemplar bounding boxes. We additionally show the learned prototype model for a sample of images from the plant dataset TPC-268~\cite{xu2026plant} in Figure~\ref{fig:tpc}.

\subsubsection{Evaluation and metrics.}
We report metrics for element modeling, image modeling, and object count. 
To assess element modeling quality, we measure how accurately a method reconstructs each of the three annotated exemplars with a single element. More precisely, for each exemplar, we select for each method the predicted element whose mask overlaps the most with the exemplar bounding box, which defines a foreground and a background mask. We then build a reconstructed exemplar by compositing the selected element with a uniform background, using as color the average ground-truth color over the background mask. Note that this does not require a ground-truth mask and avoids favoring methods that use several elements to reconstruct a single exemplar. These single-element exemplar reconstructions are also the ones we show in our qualitative results. 
To assess image modeling quality, we composite all elements predicted by a method with a uniform background and compare the result to the ground-truth image. Similar to what we do for the exemplar metric, we use as background color the mean color of the input image outside the predicted elements mask. 

For both object and image modeling, we report PSNR, SSIM~\cite{wang2004image} and LPIPS~\cite{zhang2018unreasonable}. For counting metrics, we report the standard mean average error (MAE) and root mean square error (RMSE) metrics.

\begin{table*}[t]
\small
\setlength{\tabcolsep}{4pt}
\centering
\caption{\textbf{Quantitative results.} $\uparrow$ indicates higher is better, $\downarrow$ lower is better. The {\it Ex } column indicates methods that require an exemplar or a text prompt to be provided, relying on user input at test time. The {\it Sup } column indicates methods that rely on supervised training from human annotations. 
}
\label{tab:quant_results}
\resizebox{\columnwidth}{!}{
\begin{tabular}{@{}lcccccccccc@{}}
\toprule

\multirow{2}{*}{\vspace{-0.5em}Method} & \multirow{2}{*}{Ex}& \multirow{2}{*}{Sup}
& \multicolumn{2}{c}{Counting}
& \multicolumn{3}{c}{Image Reconstruction} 
& \multicolumn{3}{c}{Prototype Quality} \\

\cmidrule(lr){4-5} \cmidrule(lr){6-8} \cmidrule(lr){9-11}

&&& MAE$\downarrow$ & RMSE$\downarrow$
& PSNR$\uparrow$ & SSIM$\uparrow$ & LPIPS$\downarrow$
& PSNR$\uparrow$ & SSIM$\uparrow$ & LPIPS$\downarrow$ \\
\midrule

ABC123~\cite{hobley2024abc}&&
& 20.92 & 28.58
& -- & -- & --
& -- & -- & -- \\

GeCo~\cite{pelhan2024novel} & &\textcolor{red}{\xmark}
& 7.34 & 13.66
& -- & -- & --
& -- & -- & -- \\

TMR~\cite{jo2025tmr}&\textcolor{red}{\xmark}&\textcolor{red}{\xmark}
& \textbf{4.11} & \textbf{6.82}
& -- & -- & --
& -- & -- & -- \\
\midrule

TMR+SAM+Average&(\textcolor{red}{\xmark})& (\textcolor{red}{\xmark})     
& -- & --
& 13.95 & 0.423 & 0.543 
& 17.44 & 0.658 & 0.464 \\

TMR+SpaceJAM~\cite{barel2025spacejamlightweightregularizationfreemethod}& (\textcolor{red}{\xmark})   & (\textcolor{red}{\xmark})
& -- & --
& 13.63 & 0.405 & 0.544 
& 17.77 & 0.674 & 0.430 \\

TMR+SAM+3D~\cite{zhang2023seeing}& (\textcolor{red}{\xmark})& (\textcolor{red}{\xmark})    
& -- & --
& -- & -- & -- 
& 15.53 & 0.568 & 0.500 \\

\rowcolor{gray!15}
Ours&&
& 14.11 & 26.93
& \textbf{23.76} & \textbf{0.737} & \textbf{0.242}
& \textbf{19.40} & \textbf{0.700} & \textbf{0.353} \\

\bottomrule
\end{tabular}
}
\end{table*}

\subsection{Element modeling baselines}
To the best of our knowledge, no stand-alone method can model repeated elements in an image, and existing methods require element bounding boxes or masks. We thus build baselines by first extracting element bounding boxes using TMR~\cite{jo2025tmr}, a state-of-the-art counting method, and optionally obtain element masks from its SAM~\cite{kirillov2023segment} decoder branch. Note that TMR requires exemplar annotations, so these baselines actually require more supervision than our unsupervised method. 
Our three baselines are as follows.
\begin{itemize}
    \item \textit{TMR + SAM + Average}:
    we define a prototype by averaging the segmented instances and their SAM masks, which can then be placed and scaled according to TMR bounding boxes to model the image.
    \item \textit{TMR + SpaceJAM~\cite{barel2025spacejamlightweightregularizationfreemethod}}: we use SpaceJAM to perform a joint alignment between all instances extracted by TMR and use the average of aligned instances as prototype appearance. SpaceJAM's mask is used for each exemplar.
    \item \textit{TMR + SAM + 3D object modeling~\cite{zhang2023seeing}}: we use~\cite{zhang2023seeing} to obtain parametric 3D models from the segmentations obtained by TMR and SAM. 
    This method is quite slow, requiring 10 hours to train for a single image, then more than half an hour per exemplar. Thus, for this baseline, we only evaluate element modeling. 
\end{itemize}


\subsection{Results}
\subsubsection{Comparisons}
\begin{figure}[t]
\centering
    \hspace*{1.35cm}
    \begin{overpic}[width=0.8\linewidth]{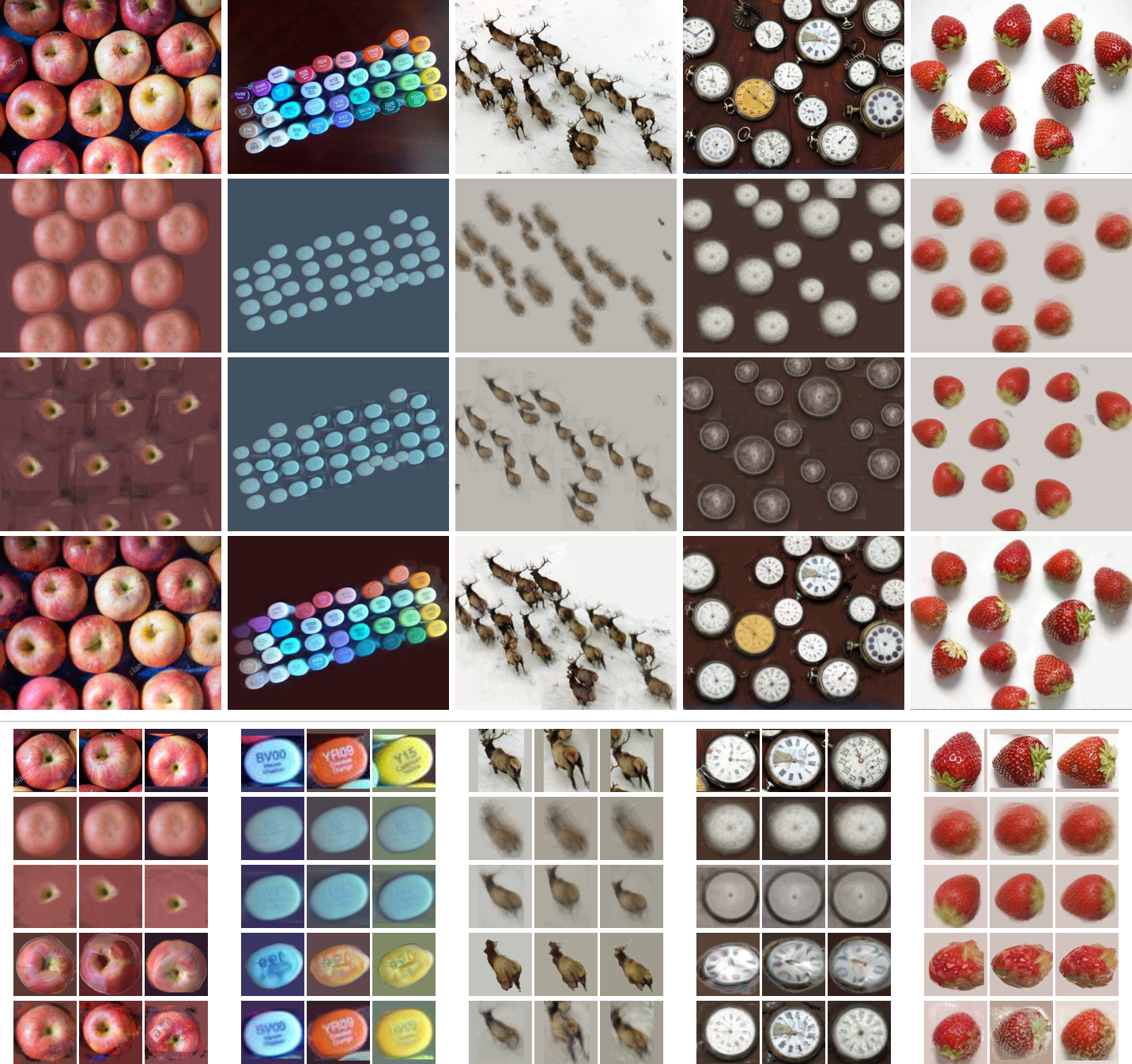}
    \put(-20.5, 86.5){\makebox(0,0)[l]{Input}}                                                   
    \put(-20.5, 70.83){\makebox(0,0)[l]{Average}}                                                
    \put(-20.5, 55.17){\makebox(0,0)[l]{SpaceJAM~\cite{barel2025spacejamlightweightregularizationfreemethod}}} 
    \put(-20.5, 39.5){\makebox(0,0)[l]{\textbf{Ours}}}                                           
    
    \put(-20.5, 27){\makebox(0,0)[l]{GT}}                                                       
    \put(-20.5, 21){\makebox(0,0)[l]{Average}}                                                  
    \put(-20.5, 15){\makebox(0,0)[l]{SpaceJAM~\cite{barel2025spacejamlightweightregularizationfreemethod}}} 
    \put(-20.5,  9){\makebox(0,0)[l]{3D model~\cite{zhang2023seeing}}}                             
    \put(-20.5,  3){\makebox(0,0)[l]{\textbf{Ours}}}                                            
  \end{overpic}
    \caption{{ \bf Qualitative comparison.} Top: Full-image reconstructions, Bottom: Exemplar reconstructions for the three exemplars annotated in the dataset. }
  \label{fig:comparison}
\end{figure}
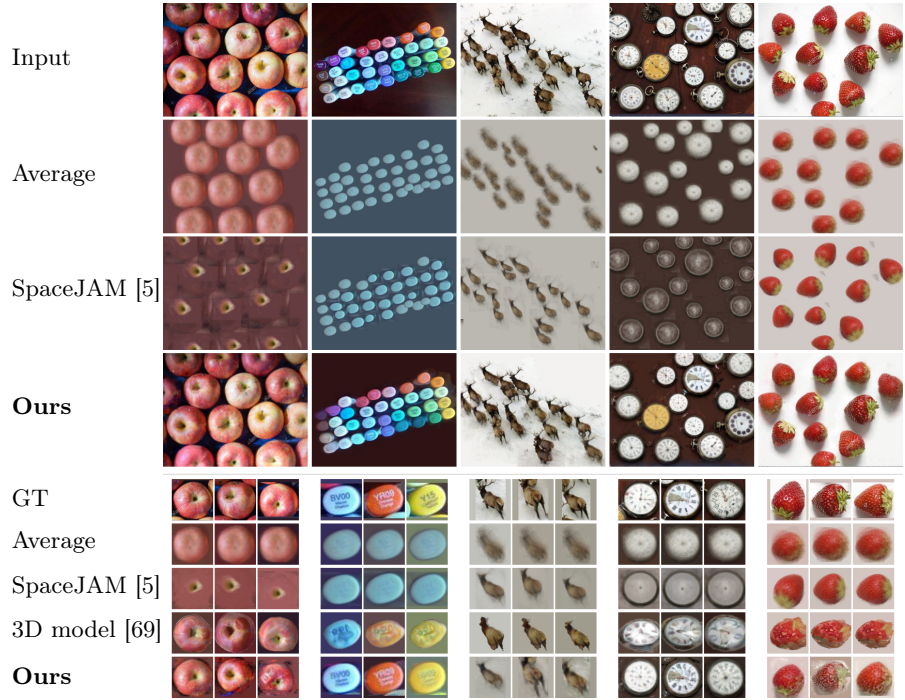

In Table~\ref{tab:quant_results}, we compare our exemplar-free approach to counting methods and element modeling baselines, and in Figure~\ref{fig:comparison} to modeling baselines.

ABC123~\cite{hobley2024abc} is an exemplar-free {multiclass} counting method, trained only on synthetic data, which reports 4 counts per image. To avoid penalizing class ambiguity (\eg splitting red and green apples), we consider, for each image, the count with minimum error between (i) each of the 4 predicted counts and (ii) the sum of the counts. To the best of our knowledge, ABC123 is the best counting method relying neither on human supervision during training nor image prompts. GeCo~\cite{pelhan2024novel} is a supervised counting network that supports exemplar-free counting via a pretrained latent prototype. TMR~\cite{jo2025tmr} is a state-of-the-art exemplar based supervised training method. Although not primarily designed for counting, our method outperforms ABC123. It is outperformed by the other supervised counting methods, which is expected. 

Our element modeling baselines are unsupervised but require access to all individual element bounding boxes or segmentations. For fair comparison, we obtain them through the best counting method, TMR, which requires both supervision and an exemplar. Still, our method outperforms all of these baselines on element and image modeling. This advantage is also evident in the qualitative comparison shown in Figure~\ref{fig:comparison}, where our method learns a meaningful element and faithfully reconstructs each of the instances. While SpaceJAM~\cite{barel2025spacejamlightweightregularizationfreemethod} performs an alignment similar to ours in spirit, its saliency mask can degenerate, in which case it models only part of the object, \eg in the apple example. It is also unable to model color or appearance variation, which is obvious in the watch example. The 3D method~\cite{zhang2023seeing} can also struggle with larger variations in appearance, and more critically, is very sensitive to the quality of its input segmentation masks. 


\subsubsection{Ablations} 
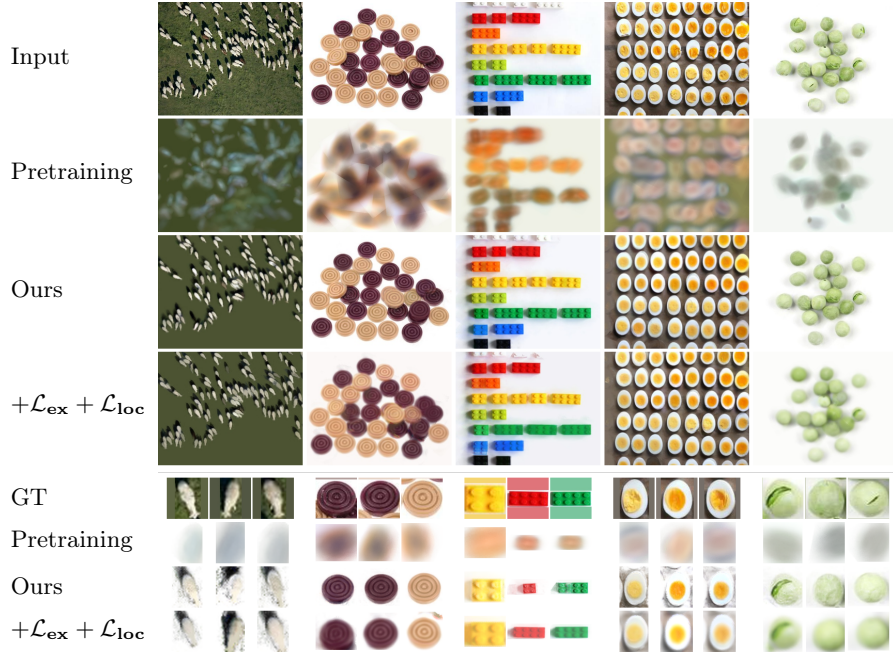
\begin{figure}[t]
\centering
    \hspace*{1.3cm}
  \begin{overpic}[width=0.8\linewidth]{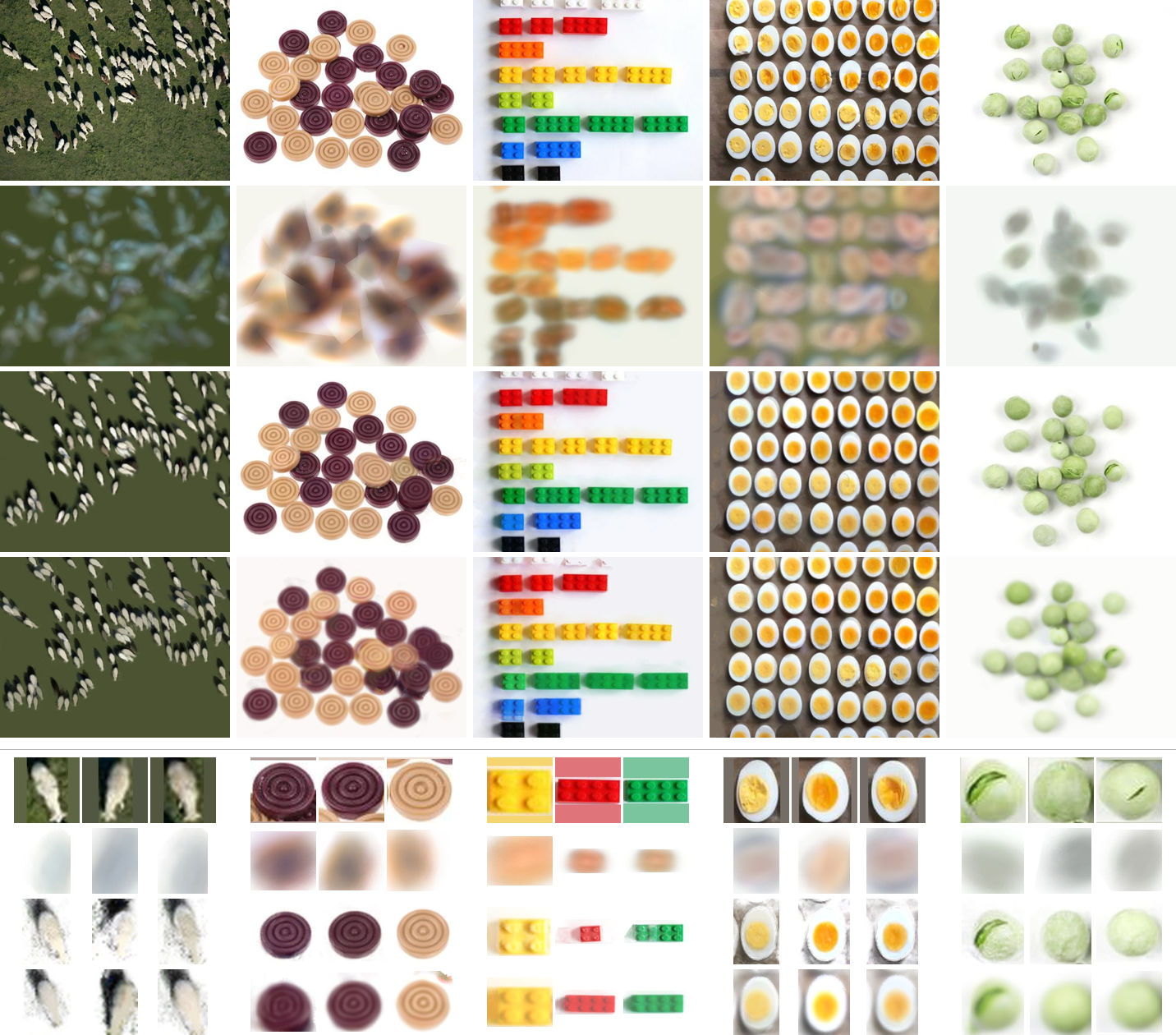}
    
    \put(-20, 80.5){\makebox(0,0)[l]{Input}}                                                   
    \put(-20, 64.83){\makebox(0,0)[l]{Pretraining}}                                                
    \put(-20, 49.17){\makebox(0,0)[l]{Ours}}                         
    \put(-20, 33.5){\makebox(0,0)[l]{\textbf{$+\mathcal{L}_\text{ex}+\mathcal{L}_\text{loc}$}}}                                           
    
    \put(-20, 21){\makebox(0,0)[l]{GT}}                                                   
    \put(-20, 15){\makebox(0,0)[l]{Pretraining}} 
    \put(-20,  9){\makebox(0,0)[l]{Ours}}
    \put(-20,  3){\makebox(0,0)[l]{\textbf{$+\mathcal{L}_\text{ex}+\mathcal{L}_\text{loc}$}}}  
    
  \end{overpic}
  \caption{\textbf{Ablations and Comparison.} Top: Full-image reconstructions, Bottom: Exemplar reconstructions for the three exemplars annotated in the dataset. }
  \label{fig:ablation}
\end{figure}

We report ablations and variants of our method in Table~\ref{tab:ablation1} and visualize some in Figure~\ref{fig:ablation}. 

Removing the appearance head, i.e., modeling each image with a single prototype, has a smaller impact on the counting metrics but a more important one on full image and prototype modeling, in particular for the more perceptual SSIM and LPIPS metrics. This is showcased in the rightmost example of Figure~\ref{fig:ablation} as well, where this variant of our method lacks the capacity to model the individual elements. Removing the sparsity loss, relying only on image reconstruction, degrades slightly not only counting accuracy but also our modeling metrics.
\paragraph{Optimization} Ablations show that all of our optimization steps are actually important for our final results. Directly optimizing full image reconstruction, without either pretraining or crop-based training leads to very bad results for all our metrics. Removing either pretraining, crop-based optimization, or image refinement leads to smaller performance drops. Pretraining is particularly critical for counting, and crop-based optimization for modeling quality, while full image refinement slightly boosts modeling.

\begin{table}[!t]
\centering
\caption{{\bf Ablations and analysis}. We put in bold any result better than our unsupervised exemplar-free method.}
\label{tab:ablation1}
\resizebox{\columnwidth}{!}{
\begin{tabular}{l cc ccc ccc}
\toprule
\multirow{2}{*}{\vspace{-0.5em}Method} & \multicolumn{2}{c}{Counting} & \multicolumn{3}{c}{Full Image} & \multicolumn{3}{c}{Prototype} \\
\cmidrule(r){2-3} \cmidrule(lr){4-6} \cmidrule(l){7-9}
& MAE$\downarrow$ & RMSE$\downarrow$ & PSNR$\uparrow$ & SSIM$\uparrow$ & LPIPS$\downarrow$ & PSNR$\uparrow$ & SSIM$\uparrow$ & LPIPS$\downarrow$ \\
\midrule
\rowcolor{gray!15}
Ours
& 14.11 & 26.93
& 23.76 & 0.737 & 0.242
& 19.40 & 0.700 & 0.353 \\

 \hspace{1em}w/o appearance head  & 14.99 &  27.31 
  & 20.58 & 0.606 & 0.384 
  & 18.53 & 0.673 & 0.432 \\
 \hspace{1em}w/o sparsity loss           & 15.62 & 28.24 & 22.57 & 0.689 & 0.298 & 18.89 & 0.691 & 0.389 \\
\multicolumn{9}{l}{\textit{Optimization ablation}} \\
\hspace{1em}direct optimization & 27.54 & 47.75 & 15.84 & 0.524 & 0.493 & 16.88 & 0.632 & 0.490\\
 \hspace{1em}w/o pretraining  & 18.81 & 38.13 & 21.08 & 0.689 & 0.330 & 18.29 & 0.655 & 0.469 \\
 \hspace{1em}w/o crop-based optimization
& 15.25 & 28.85 & 17.45 & 0.516 & 0.511 & 16.61 & 0.627 & 0.496 \\ 
 \hspace{1em}w/o image refinement
& 14.11 & 26.93 & 20.07 & 0.576 & 0.443 & 18.47 & 0.665 & 0.468 \\
\multicolumn{9}{l}{\textit{Pretraining analysis}} \\
 \hspace{1em}pretraining only 
& 15.25 & 28.85 & 14.86 & 0.396 & 0.789 & 15.48 & 0.603 & 0.659 \\
 \hspace{1em}unsup. pretraining & 15.48 & 28.63 & 22.67 & 0.666 & 0.280 & 19.21 &0.733 & 0.362 \\ 
\multicolumn{9}{l}{\textit{Leveraging annotations}} \\
 \hspace{1em}$+\mathcal{L}_\text{ex}$ & \bf 14.03 & 27.18 
  & 22.11 & 0.701 & 0.282 
  & \bf 21.54 & \bf 0.751 & \bf 0.291 \\
 \hspace{1em}$+\mathcal{L}_\text{ex}+\mathcal{L}_\text{loc}$ & \bf 7.07 & \bf 11.26 
  &  20.00 &  0.644 &  0.343 
  & \bf 20.28 & \bf 0.729 & \bf 0.330 \\
\bottomrule
\end{tabular}
}
\end{table}
\paragraph{Pretraining} Pretraining alone leads to relatively good counting metrics, but very bad modeling. This is because our pretraining model only provides very coarse localization of blob-like objects, as can be seen in Figure~\ref{fig:ablation}. Interestingly, not using bounding box and count supervision during synthetic pretraining has limited effect on the performances. This suggests that it might be possible to pretrain the model on real datasets, if they were available, which might in turn lead to better modeling results when applying the model to real images.

\paragraph{Annotations} Leveraging additional supervision from annotations naturally boosts our results. The exemplar loss has a limited impact on counting and full image modeling metrics, but strongly improves prototype modeling. Adding the localization loss strongly boosts counting performance but it hurts full image reconstruction modeling, since fewer prototypes are used to reconstruct the background. Additionally, as shown in the lego example of Figure~\ref{fig:ablation}, where the lack of supervision leads to a decomposition of the red exemplar into two pieces, this setting is particularly helpful for ambiguities due to self-similarity.

Finally, we find that using a uniform prior over object scale leads to marginal degradation in reconstruction metrics and to a degradation of counting metrics with increases of 4.53 MAE and 6.1 RMSE. On 90\% of the dataset, counting MAE increases by only 1.3, with the remaining 10\% (smallest objects) accounting for most of the gap. The scale prior is needed only in the extreme regime of very small objects.

\subsubsection{Image Manipulation.}
\renewcommand{\arraystretch}{0.8}
\begin{wrapfigure}[8]{r}{0.6\linewidth}
\vspace{-28pt}
\centering
\setlength{\tabcolsep}{2pt}
\renewcommand{\arraystretch}{0.8}
\includegraphics[width=0.8\linewidth]{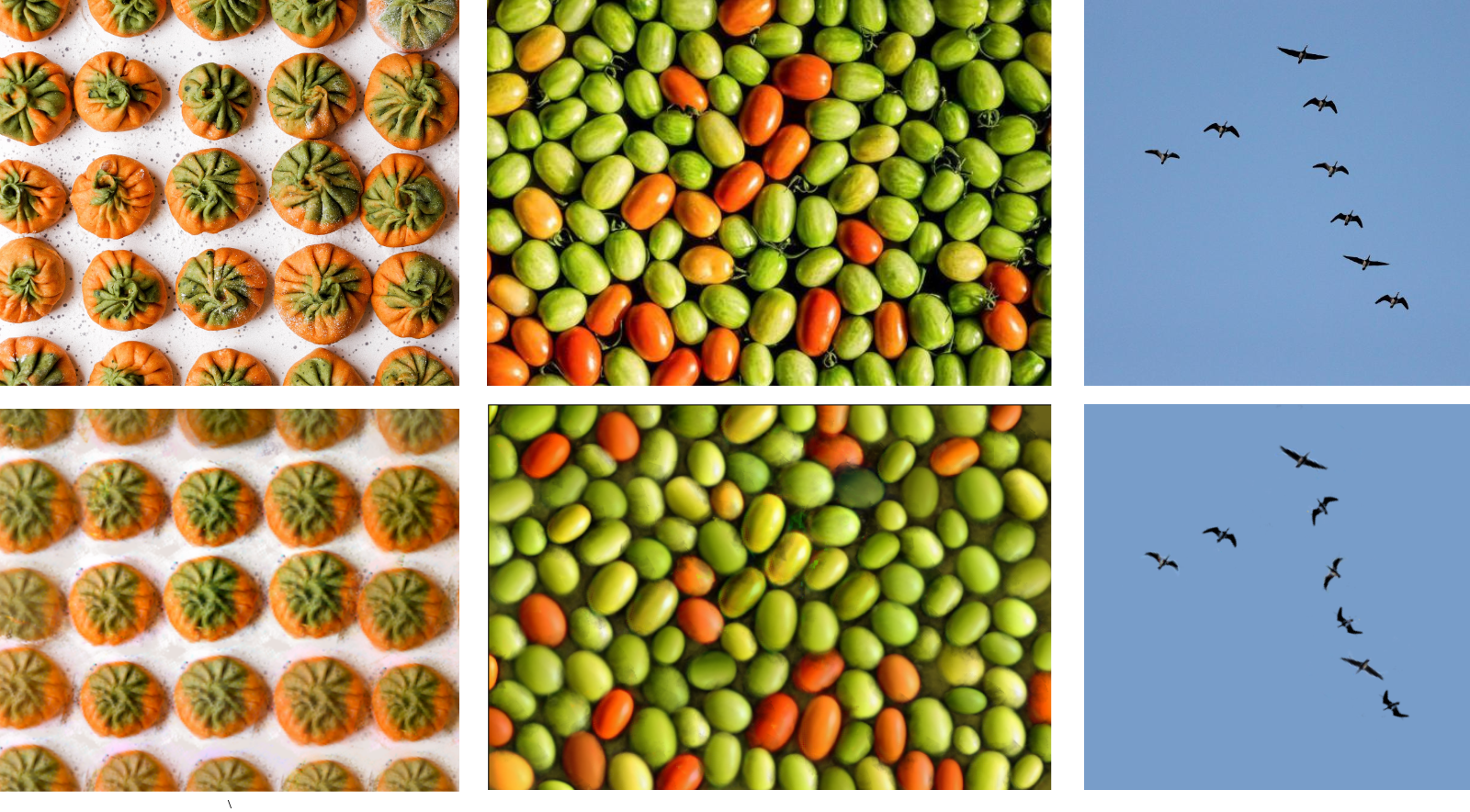}
\caption{Pose and Appearance manipulation.}
\label{fig:manip}
\end{wrapfigure}
We demonstrate our model's interpretability through image manipulation (Figure~\ref{fig:manip}; top: input, bottom: manipulation), editing the rotation, color, and appearance of individual instances.

\section{Discussion and Conclusion}
\paragraph{Limitations} Our method assumes a non-cluttered background and elements of moderate aspect ratio, and relies on a coarse scale prior, which limits its ability to handle extreme intra-image scale variation. Because grouping is driven by visual similarity in image space rather than semantics, instances that are semantically related but visually dissimilar are not recovered. Finally, part–whole ambiguity, a problem shared with counting models operating under similar conditions, persists as inherent to data with intra-image repetition.

\paragraph{Conclusion}  Revisiting a fundamental computer vision problem with modern optimization tools, we have demonstrated an analysis-by-synthesis approach that enables completely bottom-up discovery and modeling of repeated elements in real images. Our resulting image model not only improves over baselines but is particularly explainable, providing position, color, and a 1D appearance feature for each element instance. More broadly, our results suggest that meaningful object models can emerge directly from pixels. The learned decomposition also opens a promising direction for object-centric learning: intra-image repetition can serve as a free self-supervision signal.

\section*{Acknowledgments}
This work was funded by the EIDA (ANR-22-CE38-0014) and ERC DISCOVER (No. 101076028) projects, and used HPC resources from GENCI-IDRIS (No. AD011015521). It was also supported by the French government via Hi! PARIS under France 2030 (ANR-23-IACL-0005). We thank Ségolène Albouy, Raphaël Baena, Sonat Baltaçi, Paul Kervegan, Marta Lopez, Fei Meng, and Malamatenia Vlachou for figure feedback, fruitful discussions, and interesting datasets.
\bibliographystyle{splncs04}
\bibliography{main}

@String(CVPR  = {IEEE Conf. Comput. Vis. Pattern Recog.})

@String(ICCV  = {Int. Conf. Comput. Vis.})

@String(ECCV  = {Eur. Conf. Comput. Vis.})

@String(ICML  = {Int. Conf. Mach. Learn.})

@String(ICLR  = {Int. Conf. Learn. Represent.})

@String(AAAI  = {AAAI})

@String(TOG   = {ACM Trans. Graph.})

@String(CVPR  = {CVPR})

@String(ICCV  = {ICCV})

@String(ECCV  = {ECCV})

@String(ICML  = {ICML})

@String(ICLR  = {ICLR})

@String(TOG   = {ACM TOG})

@inproceedings{leung1996detecting,
  title={Detecting, localizing and grouping repeated scene elements from an image}, 
  author={Leung, Thomas and Malik, Jitendra},
  booktitle={European Conference on Computer Vision},
  pages={546--555},
  year={1996},
  organization={Springer}
}

@article{wang2004image,
  title={Image quality assessment: from error visibility to structural similarity},
  author={Wang, Zhou and Bovik, Alan C and Sheikh, Hamid R and Simoncelli, Eero P},
  journal={IEEE transactions on image processing},
  volume={13},
  number={4},
  pages={600--612},
  year={2004},
  publisher={IEEE}
}

@article{kosiorek2019stacked,
  title={Stacked capsule autoencoders},
  author={Kosiorek, Adam and Sabour, Sara and Teh, Yee Whye and Hinton, Geoffrey E},
  journal={Advances in neural information processing systems},
  volume={32},
  year={2019}
}

@article{sabour2017dynamic,
  title={Dynamic routing between capsules},
  author={Sabour, Sara and Frosst, Nicholas and Hinton, Geoffrey E},
  journal={Advances in neural information processing systems},
  volume={30},
  year={2017}
}

@inproceedings{chen2022learning, 
  title={Learning continuous implicit representation for near-periodic patterns},
  author={Chen, Bowei and Zhi, Tiancheng and Hebert, Martial and Narasimhan, Srinivasa G},
  booktitle={European Conference on Computer Vision},
  pages={529--546},
  year={2022},
  organization={Springer}
}

@incollection{schaffalitzky1999geometric,
  title={Geometric grouping of repeated elements within images},
  author={Schaffalitzky, Frederik and Zisserman, Andrew},
  booktitle={Shape, contour and grouping in computer vision},
  pages={165--181},
  year={1999},
  publisher={Springer}
}

@article{muller2007image, 
  title={Image-based procedural modeling of facades},
  author={M{\"u}ller, Pascal and Zeng, Gang and Wonka, Peter and Van Gool, Luc},
  journal={ACM Trans. Graph.},
  volume={26},
  number={3},
  pages={85},
  year={2007}
}

@inproceedings{hays2006discovering, 
  title={Discovering texture regularity as a higher-order correspondence problem},
  author={Hays, James and Leordeanu, Marius and Efros, Alexei A and Liu, Yanxi},
  booktitle={European Conference on Computer Vision},
  pages={522--535},
  year={2006},
  organization={Springer}
}

@inproceedings{ahuja2007extracting,
  title={Extracting texels in 2.1 D natural textures},
  author={Ahuja, Narendra and Todorovic, Sinisa},
  booktitle={2007 IEEE 11th International Conference on Computer Vision},
  pages={1--8},
  year={2007},
  organization={IEEE}
}

@inproceedings{doubek2010image, 
  title={Image matching and retrieval by repetitive patterns},
  author={Doubek, Petr and Matas, Jiri and Perdoch, Michal and Chum, Ondrej},
  booktitle={2010 20th international conference on pattern recognition},
  pages={3195--3198},
  year={2010}, 
  organization={IEEE}
}

@InProceedings{Liu_2015_ICCV, 
author = {Liu, Siying and Ng, Tian-Tsong and Sunkavalli, Kalyan and Do, Minh N. and Shechtman, Eli and Carr, Nathan},
title = {PatchMatch-Based Automatic Lattice Detection for Near-Regular Textures},
booktitle = {Proceedings of the IEEE International Conference on Computer Vision (ICCV)},
month = {December},
year = {2015}
}

@article{cheng2010repfinder, 
  title={Repfinder: finding approximately repeated scene elements for image editing},
  author={Cheng, Ming-Ming and Zhang, Fang-Lue and Mitra, Niloy J and Huang, Xiaolei and Hu, Shi-Min},
  journal={ACM transactions on graphics (TOG)},
  volume={29},
  number={4},
  pages={1--8},
  year={2010},
  publisher={ACM New York, NY, USA}
}

@inproceedings{violante2025splat,
  title={Splat and Replace: 3D Reconstruction with Repetitive Elements},
  author={Violante, Nicol{\'a}s and Meuleman, Andreas and Gauthier, Alban and Durand, Fredo and Groueix, Thibault and Drettakis, George},
  booktitle={Proceedings of the Special Interest Group on Computer Graphics and Interactive Techniques Conference Conference Papers},
  pages={1--12},
  year={2025}
}

@inproceedings{glasner2009super,
  title={Super-resolution from a single image},
  author={Glasner, Daniel and Bagon, Shai and Irani, Michal},
  booktitle={2009 IEEE 12th international conference on computer vision},
  pages={349--356},
  year={2009},
  organization={IEEE}
}

@inproceedings{huang2015single,
  title={Single image super-resolution from transformed self-exemplars},
  author={Huang, Jia-Bin and Singh, Abhishek and Ahuja, Narendra},
  booktitle={Proceedings of the IEEE conference on computer vision and pattern recognition},
  pages={5197--5206},
  year={2015}
}

@inproceedings{efros1999texture, 
  title={Texture synthesis by non-parametric sampling},
  author={Efros, Alexei A and Leung, Thomas K},
  booktitle={Proceedings of the seventh IEEE international conference on computer vision},
  volume={2},
  pages={1033--1038},
  year={1999},
  organization={IEEE}
}

@inproceedings{zhang2023seeing,
  title={Seeing a rose in five thousand ways},
  author={Zhang, Yunzhi and Wu, Shangzhe and Snavely, Noah and Wu, Jiajun},
  booktitle={Proceedings of the IEEE/CVF Conference on Computer Vision and Pattern Recognition},
  pages={962--971},
  year={2023}
}

@article{cheng2024structure,
  title={Structure from Duplicates: Neural Inverse Graphics from a Pile of Objects},
  author={Cheng, Tianhang and Ma, Wei-Chiu and Guan, Kaiyu and Torralba, Antonio and Wang, Shenlong},
  journal={arXiv preprint arXiv:2401.05236},
  year={2024}
}

@inproceedings{sinlayout2024,
      author    = {Zhao, Linan and Yuan, Zeqing and Zhang, Yunzhi and Wu, Shangzhe and Wu, Jiajun},
      title     = {Learning Generative 3D Scene Layouts from a Single Image},
      booktitle = {CVPR},
      series    = {AI for 3D Generation Workshop},
      year      = {2024},
    }

@inproceedings{huang2007unsupervised,
  title={Unsupervised joint alignment of complex images},
  author={Huang, Gary B and Jain, Vidit and Learned-Miller, Erik},
  booktitle={2007 IEEE 11th international conference on computer vision},
  pages={1--8},
  year={2007},
  organization={IEEE}
}

@article{learned2005data,
  title={Data driven image models through continuous joint alignment},
  author={Learned-Miller, Erik G},
  journal={IEEE Transactions on Pattern Analysis and Machine Intelligence},
  volume={28},
  number={2},
  pages={236--250},
  year={2005},
  publisher={IEEE}
}

@inproceedings{barel2025spacejamlightweightregularizationfreemethod,
  title={Spacejam: a lightweight and regularization-free method for fast joint alignment of images},
  author={Barel, Nir and Weber, Ron Shapira and Mualem, Nir and Finder, Shahaf E and Freifeld, Oren},
  booktitle={European Conference on Computer Vision},
  pages={180--197},
  year={2024},
  organization={Springer}
}

@inproceedings{ofri2023neural,
  title={Neural Congealing: Aligning Images to a Joint Semantic Atlas},
  author={Ofri-Amar, Dolev and Geyer, Michal and Kasten, Yoni and Dekel, Tali},
  booktitle=CVPR,
  year={2023}
}

@inproceedings{gupta2023asic,
  title={Asic: Aligning sparse in-the-wild image collections},
  author={Gupta, Kamal and Jampani, Varun and Esteves, Carlos and Shrivastava, Abhinav and Makadia, Ameesh and Snavely, Noah and Kar, Abhishek},
  booktitle={Proceedings of the IEEE/CVF International Conference on Computer Vision},
  pages={4134--4145},
  year={2023}
}

@inproceedings{peebles2022gan,
  title={Gan-supervised dense visual alignment},
  author={Peebles, William and Zhu, Jun-Yan and Zhang, Richard and Torralba, Antonio and Efros, Alexei A and Shechtman, Eli},
  booktitle=CVPR,
  pages={13470--13481},
  year={2022}
}

@inproceedings{mu2022coordgan, 
  title={CoordGAN: Self-Supervised Dense Correspondences Emerge from GANs},
  author={Mu, Jiteng and De Mello, Shalini and Yu, Zhiding and Vasconcelos, Nuno and Wang, Xiaolong and Kautz, Jan and Liu, Sifei},
  booktitle={Proceedings of the IEEE/CVF Conference on Computer Vision and Pattern Recognition},
  pages={10011--10020},
  year={2022}
}

@article{hirsch2025fastjam,
  title={FastJAM: a Fast Joint Alignment Model for Images},
  author={Hirsch, Omri and Weber, Ron Shapira and Ifergane, Shira and Freifeld, Oren},
  journal={arXiv preprint arXiv:2510.22842},
  year={2025}
}

@inproceedings{jo2025tmr,
  title     = {Few-Shot Pattern Detection via Template Matching and Regression},
  author    = {Jo, Eunchan and Kang, Dahyun and Kim, Sanghyun and Choi, Yunseon and Cho, Minsu},
  booktitle = {International Conference on Computer Vision (ICCV)},
  year      = {2025},
}

@article{lempitsky2010learning,
  title={Learning to count objects in images},
  author={Lempitsky, Victor and Zisserman, Andrew},
  journal={Advances in neural information processing systems},
  volume={23},
  year={2010}
}

@inproceedings{lu2018class,
  title={Class-agnostic counting},
  author={Lu, Erika and Xie, Weidi and Zisserman, Andrew},
  booktitle={Asian conference on computer vision},
  pages={669--684},
  year={2018},
  organization={Springer}
}

@inproceedings{ranjan2022exemplar,
  title={Exemplar free class agnostic counting},
  author={Ranjan, Viresh and Nguyen, Minh Hoai},
  booktitle={Proceedings of the Asian Conference on Computer Vision},
  pages={3121--3137},
  year={2022}
}

@inproceedings{ranjan2021learning,
  title={Learning to count everything},
  author={Ranjan, Viresh and Sharma, Udbhav and Nguyen, Thu and Hoai, Minh},
  booktitle={Proceedings of the IEEE/CVF conference on computer vision and pattern recognition},
  pages={3394--3403},
  year={2021}
}

@inproceedings{djukic2023low,
  title={A low-shot object counting network with iterative prototype adaptation},
  author={{\DJ}uki{\'c}, Nikola and Luke{\v{z}}i{\v{c}}, Alan and Zavrtanik, Vitjan and Kristan, Matej},
  booktitle={Proceedings of the IEEE/CVF International Conference on Computer Vision},
  pages={18872--18881},
  year={2023}
}

@inproceedings{nguyen2022few, 
  title={Few-shot object counting and detection},
  author={Nguyen, Thanh and Pham, Chau and Nguyen, Khoi and Hoai, Minh},
  booktitle={European Conference on Computer Vision},
  pages={348--365},
  year={2022},
  organization={Springer}
}

@article{pelhan2024novel, 
  title={A novel unified architecture for low-shot counting by detection and segmentation},
  author={Pelhan, Jer and Lukezic, Alan and Zavrtanik, Vitjan and Kristan, Matej},
  journal={Advances in Neural Information Processing Systems},
  volume={37},
  pages={66260--66282},
  year={2024}
}

@article{amini2024countgd,
  title={Countgd: Multi-modal open-world counting},
  author={Amini-Naieni, Niki and Han, Tengda and Zisserman, Andrew},
  journal={Advances in Neural Information Processing Systems},
  volume={37},
  pages={48810--48837},
  year={2024}
}

@article{liu2022countr,
  title={Countr: Transformer-based generalised visual counting},
  author={Liu, Chang and Zhong, Yujie and Zisserman, Andrew and Xie, Weidi},
  journal={arXiv preprint arXiv:2208.13721},
  year={2022}
}

@inproceedings{hobley2024abc, 
  title={Abc easy as 123: A blind counter for exemplar-free multi-class class-agnostic counting},
  author={Hobley, Michael and Prisacariu, Victor},
  booktitle={European Conference on Computer Vision},
  pages={304--319},
  year={2024},
  organization={Springer}
}

@article{amini2023open,
  title={Open-world text-specified object counting},
  author={Amini-Naieni, Niki and Amini-Naieni, Kiana and Han, Tengda and Zisserman, Andrew},
  journal={arXiv preprint arXiv:2306.01851},
  year={2023}
}

@inproceedings{knobel2024learning, 
  title={Learning to count without annotations},
  author={Knobel, Lukas and Han, Tengda and Asano, Yuki M},
  booktitle={Proceedings of the IEEE/CVF Conference on Computer Vision and Pattern Recognition},
  pages={22924--22934},
  year={2024}
}

@inproceedings{huang2024point,
  title={Point segment and count: A generalized framework for object counting},
  author={Huang, Zhizhong and Dai, Mingliang and Zhang, Yi and Zhang, Junping and Shan, Hongming},
  booktitle={Proceedings of the IEEE/CVF conference on computer vision and pattern recognition},
  pages={17067--17076},
  year={2024}
}

@inproceedings{zhang2018unreasonable,
  title={The unreasonable effectiveness of deep features as a perceptual metric},
  author={Zhang, Richard and Isola, Phillip and Efros, Alexei A and Shechtman, Eli and Wang, Oliver},
  booktitle={Proceedings of the IEEE conference on computer vision and pattern recognition},
  pages={586--595},
  year={2018}
}

@article{villa2024unsupervised,
  title={Unsupervised object discovery: A comprehensive survey and unified taxonomy},
  author={Villa-V{\'a}squez, Jos{\'e}-Fabian and Pedersoli, Marco},
  journal={arXiv preprint arXiv:2411.00868},
  year={2024}
}

@article{karazija2021clevrtex,
  title={Clevrtex: A texture-rich benchmark for unsupervised multi-object segmentation},
  author={Karazija, Laurynas and Laina, Iro and Rupprecht, Christian},
  journal={arXiv preprint arXiv:2111.10265},
  year={2021}
}

@article{smirnov2021marionette,
  title={Marionette: Self-supervised sprite learning},
  author={Smirnov, Dmitriy and Gharbi, Michael and Fisher, Matthew and Guizilini, Vitor and Efros, Alexei and Solomon, Justin M},
  journal={Advances in Neural Information Processing Systems},
  volume={34},
  pages={5494--5505},
  year={2021}
}

@article{eslami2016attend,
  title={Attend, infer, repeat: Fast scene understanding with generative models},
  author={Eslami, SM and Heess, Nicolas and Weber, Theophane and Tassa, Yuval and Szepesvari, David and Hinton, Geoffrey E and others},
  journal={Advances in neural information processing systems},
  volume={29},
  year={2016}
}

@inproceedings{jia2023improving,
	author = {Jia, Baoxiong and Liu, Yu and Huang, Siyuan},
	booktitle = {International {Conference} on {Learning} {Representations} ({ICLR})},
	year = {2023},
	pages = {},
	organization = {},
	title = {Improving {Object}-centric {Learning} with {Query} {Optimization}.},
	volume = {},
}

@article{singh2021illiterate,
  title={Illiterate dall-e learns to compose},
  author={Singh, Gautam and Deng, Fei and Ahn, Sungjin},
  journal={arXiv preprint arXiv:2110.11405},
  year={2021}
}

@article{seitzer2022bridging,
  title={Bridging the gap to real-world object-centric learning},
  author={Seitzer, Maximilian and Horn, Max and Zadaianchuk, Andrii and Zietlow, Dominik and Xiao, Tianjun and Simon-Gabriel, Carl-Johann and He, Tong and Zhang, Zheng and Sch{\"o}lkopf, Bernhard and Brox, Thomas and others},
  journal={arXiv preprint arXiv:2209.14860},
  year={2022}
}

@inproceedings{monnier2021dtisprites,
title={{Unsupervised Layered Image Decomposition into Object Prototypes}},
author={Monnier, Tom and Vincent, Elliot and Ponce, Jean and Aubry, Mathieu},
booktitle={ICCV},
year={2021},
}

@inproceedings{crawford2019spair,
  title={Spatially Invariant Unsupervised Object Detection with Convolutional Neural Networks},
  author={Eric Crawford and Joelle Pineau},
  booktitle={AAAI},
  year={2019}
}

@article{lin2020space,
  title={Space: Unsupervised object-oriented scene representation via spatial attention and decomposition},
  author={Lin, Zhixuan and Wu, Yi-Fu and Peri, Skand Vishwanath and Sun, Weihao and Singh, Gautam and Deng, Fei and Jiang, Jindong and Ahn, Sungjin},
  journal={arXiv preprint arXiv:2001.02407},
  year={2020}
}

@inproceedings{greff2019multi,
  title={Multi-object representation learning with iterative variational inference},
  author={Greff, Klaus and Kaufman, Rapha{\"e}l Lopez and Kabra, Rishabh and Watters, Nick and Burgess, Christopher and Zoran, Daniel and Matthey, Loic and Botvinick, Matthew and Lerchner, Alexander},
  booktitle={International conference on machine learning},
  pages={2424--2433},
  year={2019},
  organization={PMLR}
}

@article{burgess2019monet,
  title={Monet: Unsupervised scene decomposition and representation},
  author={Burgess, Christopher P and Matthey, Loic and Watters, Nicholas and Kabra, Rishabh and Higgins, Irina and Botvinick, Matt and Lerchner, Alexander},
  journal={arXiv preprint arXiv:1901.11390},
  year={2019}
}

@article{locatello2020object,
  title={Object-centric learning with slot attention},
  author={Locatello, Francesco and Weissenborn, Dirk and Unterthiner, Thomas and Mahendran, Aravindh and Heigold, Georg and Uszkoreit, Jakob and Dosovitskiy, Alexey and Kipf, Thomas},
  journal={Advances in neural information processing systems},
  volume={33},
  pages={11525--11538},
  year={2020}
}

@article{biza2023invariant,
  title={Invariant slot attention: Object discovery with slot-centric reference frames},
  author={Biza, Ondrej and Van Steenkiste, Sjoerd and Sajjadi, Mehdi SM and Elsayed, Gamaleldin F and Mahendran, Aravindh and Kipf, Thomas},
  journal={arXiv preprint arXiv:2302.04973},
  year={2023}
}

@inproceedings{kirillov2023segment,
  title={Segment anything},
  author={Kirillov, Alexander and Mintun, Eric and Ravi, Nikhila and Mao, Hanzi and Rolland, Chloe and Gustafson, Laura and Xiao, Tete and Whitehead, Spencer and Berg, Alexander C and Lo, Wan-Yen and others},
  booktitle={Proceedings of the IEEE/CVF international conference on computer vision},
  pages={4015--4026},
  year={2023}
}

@article{simeoni2025dinov3,
  title={Dinov3},
  author={Sim{\'e}oni, Oriane and Vo, Huy V and Seitzer, Maximilian and Baldassarre, Federico and Oquab, Maxime and Jose, Cijo and Khalidov, Vasil and Szafraniec, Marc and Yi, Seungeun and Ramamonjisoa, Micha{\"e}l and others},
  journal={arXiv preprint arXiv:2508.10104},
  year={2025}
}

@article{jaderberg2015spatial,
  title={Spatial transformer networks},
  author={Jaderberg, Max and Simonyan, Karen and Zisserman, Andrew and others},
  journal={Advances in neural information processing systems},
  volume={28},
  year={2015}
}

@inproceedings{Amir:2021:VITdescriptoers,
  title={Deep vit features as dense visual descriptors},
  author={Amir, Shir and Gandelsman, Yossi and Bagon, Shai and Dekel, Tali},
  booktitle={ECCV Workshops},
  year={2022}
}

@inproceedings{caron2021emerging,
  title={Emerging properties in self-supervised vision transformers},
  author={Caron, Mathilde and Touvron, Hugo and Misra, Ishan and J{\'e}gou, Herv{\'e} and Mairal, Julien and Bojanowski, Piotr and Joulin, Armand},
  booktitle={Proceedings of the IEEE/CVF international conference on computer vision},
  pages={9650--9660},
  year={2021}
}

@inproceedings{carion2020end,
  title={End-to-end object detection with transformers},
  author={Carion, Nicolas and Massa, Francisco and Synnaeve, Gabriel and Usunier, Nicolas and Kirillov, Alexander and Zagoruyko, Sergey},
  booktitle={European conference on computer vision},
  pages={213--229},
  year={2020},
  organization={Springer}
}

@inproceedings{lin2014microsoft,
  title={Microsoft coco: Common objects in context},
  author={Lin, Tsung-Yi and Maire, Michael and Belongie, Serge and Hays, James and Perona, Pietro and Ramanan, Deva and Doll{\'a}r, Piotr and Zitnick, C Lawrence},
  booktitle={European conference on computer vision},
  pages={740--755},
  year={2014},
  organization={Springer}
}

@inproceedings{gadelha20173d,
  title={3d shape induction from 2d views of multiple objects},
  author={Gadelha, Matheus and Maji, Subhransu and Wang, Rui},
  booktitle={2017 international conference on 3d vision (3DV)},
  pages={402--411},
  year={2017},
  organization={IEEE}
}

@inproceedings{shaham2019singan,
  title={Singan: Learning a generative model from a single natural image},
  author={Shaham, Tamar Rott and Dekel, Tali and Michaeli, Tomer},
  booktitle={Proceedings of the IEEE/CVF international conference on computer vision},
  pages={4570--4580},
  year={2019}
}

@article{wang2025sindiffusion,
  title={Sindiffusion: Learning a diffusion model from a single natural image},
  author={Wang, Weilun and Bao, Jianmin and Zhou, Wengang and Chen, Dongdong and Chen, Dong and Yuan, Lu and Li, Houqiang},
  journal={IEEE Transactions on Pattern Analysis and Machine Intelligence},
  volume={47},
  number={5},
  pages={3412--3423},
  year={2025},
  publisher={IEEE}
}

@inproceedings{shocher_ingan_2019,
  title={{InGAN}: Capturing and remapping the ``{DNA}'' of a natural image},
  author={Shocher, Assaf and Bagon, Shai and Isola, Phillip and Irani, Michal},
  booktitle={Proc. ICCV},
  year={2019}
}

@inproceedings{kulikov_sinddm_2022,
  title={{SinDDM}: A single image denoising diffusion model},
  author={Kulikov, Vladimir and Yadin, Shahar and Kleiner, Matan and Michaeli, Tomer},
  booktitle={Proc. ICML},
  year={2023}
}

@inproceedings{xu2026plant,
  title={Plant Taxonomy Meets Plant Counting: A Fine-Grained, Taxonomic Dataset for Counting Hundreds of Plant Species},
  author={Xu, Jinyu and Hu, Tianqi and Hu, Xiaonan and Zhou, Letian and Cao, Songliang and Zhang, Meng and Lu, Hao},
  booktitle={Proceedings of the IEEE/CVF Conference on Computer Vision and Pattern Recognition},
  pages={167--177},
  year={2026}
}

@inproceedings{pelhan2024dave,
  title={Dave--a detect-and-verify paradigm for low-shot counting},
  author={Pelhan, Jer and Luke{\v{z}}i{\v{c}}, Alan and Zavrtanik, Vitjan and Kristan, Matej},
  booktitle={2024 IEEE/CVF Conference on Computer Vision and Pattern Recognition (CVPR)},
  pages={23293--23302},
  year={2024},
  organization={IEEE}
}
\end{document}